# Credit risk prediction in an imbalanced social lending environment

**Anahita Namvar[1,2], Mohammad Siami[2], Fethi Rabhi[1], Mohsen Naderpour[2]**

[1] *FinanceIT Research Group, University of New South Wales,*
*Sydney, NSW, Australia*
*E-mail: Anahita.namvar@gmail.com*
*f.rabhi@unsw.edu.au*

[2] *Centre for Artificial Intelligence, University of Technology Sydney,*
*Sydney, NSW, Australia*
*E-mail: Mohammad.siaminamini@uts.edu.au*
*Mohsen.Naderpour@uts.edu.au*

**Abstract**

Credit risk prediction is an effective way of evaluating whether a potential borrower will repay a loan, particularly in peer-to-peer lending where class imbalance problems are prevalent. However, few credit risk prediction models for social lending consider imbalanced data and, further, the best resampling technique to use with imbalanced data is still controversial. In an attempt to address these problems, this paper presents an empirical comparison of various combinations of classifiers and resampling techniques within a novel risk assessment methodology that incorporates imbalanced data. The credit predictions from each combination are evaluated with a G-mean measure to avoid bias towards the majority class, which has not been considered in similar studies. The results reveal that combining random forest and random under-sampling may be an effective strategy for calculating the credit risk associated with loan applicants in social lending markets.

## 1. Introduction

Social lending, also known as peer-to-peer (P2P) lending, uses online trading platforms as a channel for lending money without the interference of traditional financial intermediaries, such as banks. Conducting business on peer platforms has recently become popular because it not only reduces financing costs but also has the potential for higher profitability for both lenders and borrowers [1]. Borrowers benefit from lower interest rates; lenders receive a higher return than they would from a bank [2].

However, evaluating the creditworthiness of loan applicants is a common challenge in micro-financing, where loans are typically unsecured [2]. Further, P2P lending usually occurs in settings with a high level of information asymmetry – that is, settings where the lenders do not have complete information about the borrowers' credit history. Even when that information is available, lenders may not know how to extract useful knowledge from the data [1], and manually assessing a borrower's credit risk is rarely a practical alternative, given the high level of expertise that requires.

However, supporting collateral, certified accounts, and regular reports are available through traditional banks, which could be used to supplement credit risk prediction in P2P lending markets if doing so did not increase transaction costs [3]. Therefore, predicting a borrower's creditworthiness to support decision making on whether or not to fund particular loans has emerged as a critical problem for P2P lending platforms. More effective loan evaluation tools are needed for these lending platforms to increase their market share in the financial industry [2, 4].

Traditional loan evaluation techniques assume a balanced distribution of misclassifications; however, an imbalanced dataset is far more typical of social lending platforms. To the best of our knowledge, none of the contemporary studies into P2P lending have explored resampling approaches for class imbalance issues in credit risk prediction.

Class imbalance problems arise when there are a far greater or fewer number of objects in one class than another. Effectively predicting credit risk from an imbalanced dataset is difficult because imbalanced data affects the ability of the model to discriminate between good borrowers and potential defaulters [4],and Data mining algorithms ignore the minority classes and focus on the majority class[5].

Therefore, to increase the reliability of credit risk prediction in social lending, we aimed to study the advantages and disadvantages of various strategies for processing imbalanced data using machine learning techniques. The research in this paper makes several contributions to the literature in the new and fast-growing field of P2P lending. First, we develop a new credit risk prediction process based on computational intelligence methods, and apply the most recent dataset of lending club, one of the biggest online P2P lending platforms. To the best of our knowledge, no study has used the most recent dataset of this platform. Second, this paper introduces a new attribute we developed that helps to capture a borrower's creditworthiness. Third, we address the imbalance problem by comparing various resampling approaches to determine which ones improve creditworthiness evaluations and how. Further, we explore various machine learning classifiers and resampling techniques to determine which combinations handle imbalanced data most efficiently.

The rest of the paper is organized as follows. Section 2 provides a literature review of the loan evaluation techniques in P2P lending markets and research related to class imbalance problems. Section 3 describes our research methodology, Section 4 presents the experimental results, and Section 5 provides our conclusions and future research directions.

## 2. Literature Review

### *2.1. Loan Evaluation in P2P lending*

P2P lending has emerged as a new e-commerce platform in the financial marketplace. As with many crowd-sourced services, P2P lending is bringing new economic efficiencies to financing [1]. Currently, scholars are undertaking three streams of research into this new business model: the reasons behind the development of P2P lending (which is outside the scope of this paper); the factors and methods that affect the likelihood of defaults or funding success; and the performance of a range of P2P platforms and credit risk prediction tools for evaluating loans [3].

Some of the studies that have examined the factors affecting funding success and the risk of defaults find that strong social networking relationships, personal characteristics and variables such as the amount of the loan, the interest rate, and the term of the loan are an important factors in determining a borrower's credit risk and influence funding success[3].

However, as a new trend in finance, few studies have explored credit risk prediction in P2P lending, although this number is growing. Emekter, Tu, Jirasakuldech and Lu [3] applied logistic regression to investigate the probability of defaults and find some associations between loan defaults and credit scores, debt-to-income ratios, FICO scores, and revolving credit lines. Malekipirbazari and Aksakalli [2] used a range of machine learning methods to classify good and bad loans, such as random forest, logistic regression, k-nearest neighbor, and support vector machines. They find that using a machine learning approach is much more effective than relying on the existing financial metrics, like FICO and LC grades, which the Lending Club provides to help lenders to make loan investment decisions.

Several decision support systems based on credit scoring models have been developed to help banks decide whether or not to extend credit to loan applicants. There are two types of predictive models in credit scoring: statistical approaches and artificial intelligence methods. Statistical methods include linear discriminant analysis and logistic regression. The artificial intelligence methods include decision trees, random forest, support vector machine, neural networks, and naïve Bayes. Despite the development of more advanced AI methods for evaluating the credit risk of borrowers, simple statistical approaches such as linear discriminate analysis and logistic regression still remain popular because of their high accuracy and ease of implementation [6-9].

*2.2. Class Imbalance Problem*

The class imbalance problems are common in classification problems. Class imbalance occurs when the number of instances in one class is vastly different than the instances in another. In such cases, classifiers tend to be biased towards the majority class, while the minority class is ignored [10]. Many algorithms designed to address imbalanced data classification problems have emerged over the past decade, and resampling is one of the most important strategies for solving this issue [11]. Resampling generates a balance training dataset prior to building the classification model.

The three types of resampling techniques are over-sampling, under-sampling, and a hybrid of the two [11].

Over-sampling creates new samples in the minority class. Three prominent oversampling methods are random over-sampling, synthetic minority over-sampling technique (SMOTE), and adaptive synthetic sampling. Random over-sampling randomly duplicates the minority samples to balance the distribution of data [12]. SMOTE uses k-nearest neighbors to produce new instances based on the distance between the minority data and some randomly selected nearest neighbors [13]. Adaptive synthetic sampling uses the density distribution to generate synthetic samples for each minority instance [14].

Under-sampling discards samples from majority class [15]. Random under-sampling and instance hardness threshold are the state-of-the-art under-sampling techniques. Random under-sampling randomly eliminates examples from the majority class to balance the class distribution [12]. The instance hardness threshold method measures the probability that an instance will be misclassified and uses a constant threshold to filter instances in all iterations [16].

The third resampling approach is a hybrid method that combines over- and under-sampling. SMOTE + Tomek links (SMOTE-TOMEK) and SMOTE+ edited nearest neighbor (SMOTE-ENN) are the two hybrid techniques compared in this study. SMOTE-TOMEK corrects SMOTE data by finding pairs of minimally distanced nearest neighbors of opposite classes. It then identifies and removes Tomek links to produce a balanced dataset with well-defined classes. SMOTE-ENN follows the same procedure as SMOTE-Tomek but uses edited nearest neighbors to balance the dataset.

Class imbalance is a common problem in loan default prediction. Abeysinghe, Li and He [17], Brown and Mues [18], and Sanz, Bernardo, Herrera, Bustince and Hagras [19] have all provided solutions to class imbalance problems for credit risk prediction. However, to the best of our knowledge, only a few studies have addressed the class imbalance problem in detail in the social lending context.

3. **Research Methodology**

To help lenders evaluate the creditworthiness of borrowers in social lending platforms, we developed a decision support system that includes a novel prediction model to reduce the risk of loan defaults. Figure 1 illustrates the model; The model first takes a raw data and passed it through feature engineering step to clean data and select the appropriate number of features. After that, the prepared data is divided into the training set and testing set. The challenge is that the imbalanced dataset is a common problem in credit risk evaluation, and it can cause misclassification. So, resampling approaches have taken into consideration to solve imbalance issue in training set, then balanced data feed to state of the art prediction models to train the model. Then the classification results are validated, we may feed back and revise our approach to reaching better classification results.

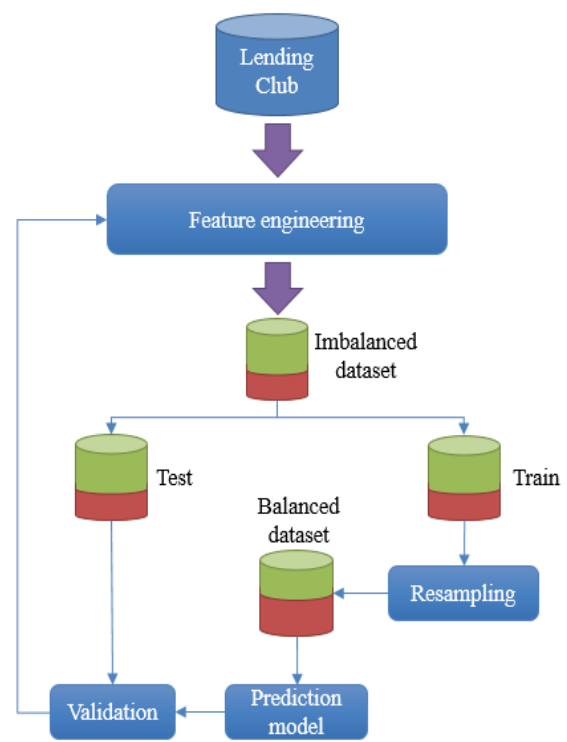

Fig. 1. Research methodology

Since this study intends to investigate the most efficient combination of resampling approaches and state of the art machine learning algorithms; therefore, the model implementation will be repeated for different combinations.

In the following subsections, each model component is explained in detail.

### 3.1. Feature Engineering

The first component in the model is a feature engineering module. Its main purpose is to enhance data reliability by cleaning data and selecting the subset of data features with the most discriminatory power. In credit risk prediction, ignoring irrelevant features can increase classification accuracy [20] and decrease the computational costs associated with running several machine learning models [21]. Feature selection also reduces the dimensionality of the data, which helps mitigate the risk of overfitting. Our model comprises four important steps in feature engineering: data cleaning, leaky data removal, data transformation and correlation analysis, and deriving new attributes.

The data is cleaned by first removing missing and null values from dataset [22], then all outliers are removed according to the acceptable range defined in [23]. Eq. (1) shows the upper bound and lower bounds of this range.

$$Lower\ bound < Acceptable\ range < Upper\ bound \quad (1)$$

$Lower\ band = [Q1 - 1.5 \times (Q3 - Q1)]$
$Upper\ band = [Q3 - 1.5 \times (Q3 - Q1)]$

Where Q3 is the third quartile and Q1 is the first quartile.

Once the data has been cleaned, the features that may cause data leakage in the model are identified. These features typically do not have available values at the time the prediction model is used. So, training model with these features may make unrealistically good predictions. Surprisingly, the Lending Club dataset contains leakage data, but few studies relying on this dataset have taken that into consideration.

Data transformation including; converting categorical features to numeric, standardization and log transformation. Some classification algorithms, such as logistic regression, cannot handle categorical features, so they are transformed into new forms of data. For standardization, the min-max normalization in Eq. (2) ensures that all parameters use the same scale.

$$X_n = \frac{X - X_{min}}{X_{max} - X_{min}} \quad (2)$$

A log transformation reduces skewness in the data distribution (which cannot be applied to zero or negative values).

Correlations for numeric and binarized nominal attributes are then computed with respect to the loan status to provide a better understanding of the data and its attributes. Finally, according to Malekipirbazari and Aksakalli [2], we define variables that are simple ratio of other features. Moreover, this study introduces the new non-standard financial feature. These ratios help to capture certain borrower characteristics to make the most of the available data.

### 3.2. Imbalanced Learning Approaches

This study employs resampling approach to deal with the imbalance problem. Figure2 demonstrates three categories of resampling approach including under-sampling, over-sampling, and hybrid methods. We address the state of the art algorithms in each of these categories. The under-sampling approach includes random under-sampling (RUS), and instance hardness threshold(IHT) algorithms. For over-sampling approach, random over-sampling (ROS), synthetic minority over-sampling technique (SMOTE), and adaptive synthetic sampling (ADASYN) are studied. Finally, SMOTE + Tomek links (SMOTE-TOMEK) and SMOTE+ edited nearest neighbor (SMOTE-ENN) are two prominent hybrid approaches that considered by this research. Section 2.2. presents complete literature on these algorithms.

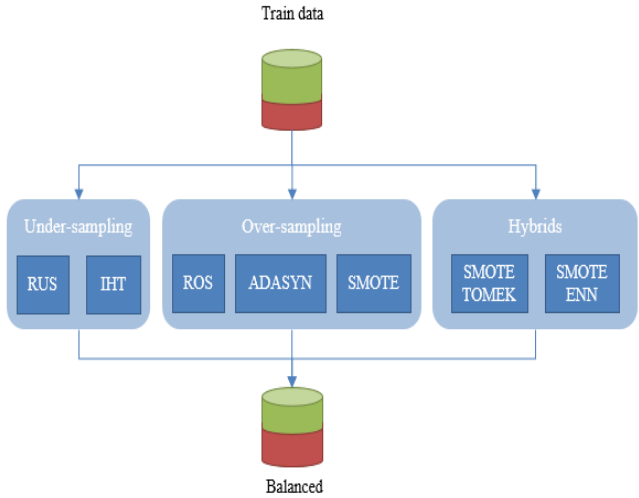

Fig. 2. Imbalanced learning approach

### 3.3. Classification Models

To the best of our knowledge, logistic regression, linear discriminate analysis, and random forest have demonstrated the best performance in the area of classification. Therefore, these three algorithms were selected for loans evaluation in this research.

#### 3.1.1. Logistic Regression

Logistic regression is a standard industry algorithm that is commonly used in practice because of its simplicity and balanced error distribution [24]. It is a binary classification technique that generates one of two variables as its result, e.g., good or bad borrowers. The logistic regression formula is shown in Eq. (3).

$$Ln\left(\frac{F(x)}{1-F(x)}\right) = \beta_0 + \sum_{i=1}^{n}\beta_i X_i \quad (3)$$

Where F(x) refers to the probability prediction, $\beta_0$ is the constant coefficient, and $\beta_i$ is the coefficient for the feature $x_i$, which is calculated using maximum likelihood. Therefore, for a set of features, $x_i$ $i=1,...,n$, the logistic regression algorithm predicts the probability that a sample belongs to a specific class [2, 24, 25].

#### 3.1.2. Linear Discriminate Analysis

Linear discriminate analysis is a statistical algorithm that determines the relationship between a target variable and a set of independent variables [26]. Many studies into credit scoring have used linear discriminate analysis because it tends to achieve better performance than other classifiers when linear patterns are involved [27].

#### 3.1.3. Random Forest

Random forest algorithms are based on ensemble trees. This method, which can be seen as an enhanced bagging technique, is a powerful way to construct a forest of random decision trees. A random forest algorithm can also build multiple decision trees that have been trained on bootstrap samples from the training data. Rather than considering all available features, the algorithm randomly chooses a subset of attributes when building the trees or splitting the nodes. Once all the trees have been generated, the most popular class is decided with a voting function [2, 6, 24].

### 3.4. Validation

The dataset was divided into a train set and a test set at a ratio of 70:30. Only the training set was balanced, through resampling, then validated with the still imbalanced test set.

### 4. Experimental Results

All experiments were conducted in Python (version 2.7.13), on MacBook Pro with 2GHz Intel Core i7 CPU, and 8GB of memory. The average values of 20 repetitions of learning procedure with random numbers are reported [15].

### 4.1. Dataset Description

Advances in P2P lending markets have generated large amounts of data on real-world P2P lending transactions. This analysis is based on 2016 and 2017 data from the publicly available datasets released by the Lending Club, a well-known P2P lending platform (lendingclub.com).

The dataset contains approximately 636K borrower records and 145 features in total. Redundant information, noise, and leakage features were removed from the dataset. The features detected as leaks were: LC grade, interest rate, issue date, outstanding principal, total payment, total received principle, total received interest, total late fees received, recoveries, post charge off collection fee, last payment date, last payment amount, and fund amount.

The variables with skewed distributions were annual income, income to payment ratio, and revolving to income ratio. Log transformations were applied to create normal distributions for these variables.

Four categorical attributes were selected as nominal attributes and transformed into binarized data: term (2 categories), home ownership (3 categories), verification status (3 categories), and purpose (12 categories). Ultimately, 2+3+3+12 = 20 numerical attributes replaced these four categorical attributes.

The new attributes, defined based on monthly income, installment amount, and revolving amount features were Income-to payment ratio and revolving-to-payment ratio [2]. Further, we defined a new ratio that, to the best of our knowledge, has not been previously used in any study called new debt to income (New DTI). New DTI considers the impact of the loan repayments, should it be granted, on the borrower's solvency. It is a ratio of the repayment amount to the borrower's monthly income if the loan is approved and is defined as:

$$New\ DTI = \frac{New\ Monthly\ Repayment\ Amount\ (NMRA)}{Monthly\ income} \quad (4)$$

$NMRA = DTI * (Annual\ Income/12) + installment$

Where DTI, annual income, and installment amount are Lending Club features.

The features used in the credit risk prediction process are presented in Table 1, grouped into three categories: loan characteristics, creditworthiness, and solvency [4].

For a better understanding of the variables, we also performed a correlation analysis of all the variables with respect to loan status. Table 2 shows the correlations for the top-20 attributes.

LC grade and interest rate sit at the top of the list with the highest correlations to loan status. The Lending Club assigns an LC grade of between A and G, where A represents safer loans and G represents riskier loans. These grades were transformed into numbers between 1 and 7 (1=A and 7=G). They also assign an interest rate to each loan. G-grade loans are given the highest interest rate; A-grade loans have the lowest rate. However, LC grades and interest rates are leaky data, so despite their high correlation with loan status, they should not be included in the modeling process or overly optimistic predictions may result.

The new DTI ratio sits in third place with a correlation of 0.171, which shows that it can have a positive effect on the classification results.

Table 1. Lending Club attributes

| Category | Attribute | Description |
|---|---|---|
| Target variable | Loan status | Current status of the loan |
| loan characteristic | Term | The number of monthly payments on the loan – either 36 or 60. |
| | Loan amount | The total amount of the loan |
| | purpose | A category provided by the borrower for the loan request. 12 purposes included. |
| Borrowers' solvency | New debt to income (New DTI) | The ratio of new monthly repayment amount (if the loan is approved) to monthly income. This considers the impact of the new loan repayments. |
| | Income to Payment Ratio | (annual income / 12) / installment |
| | Debt to income (DTI) | Ratio of the borrower's total monthly debt payments to the borrower's monthly income. |
| | home ownership | The home ownership status provided by the borrower during registration. Values are RENT, OWN, and MORTGAGE |
| | verification status | Verified income (whether or not pay slips or a bank statement have been verified by the Lending Club) values are verified, not verified, and source verified |
| | Annual income | Self-reported annual income provided by the borrower during registration. |
| | Employ Length | Employment length in years. Possible values are between 0 and 10, where 0 means less than one year and 10 means ten or more years |
| | Revolving to income Ratio | Ratio of revolving credit balance to the borrower's monthly income |
| Borrowers' creditworthiness | Revolving utilization rate | The amount of credit the borrower is using relative to all available revolving credit. (Drawn amount over the total limit) |
| | percentBcGt75 | Percentage of all bankard accounts > 75% of limit |
| | Average Current Balance | Average current balance of all accounts |
| | Total Current Balance | Total current balance of all accounts |
| | installment | The monthly payment owed by the borrower if the loan originates. |
| | inquiries last 6 months | The number of inquiries in past 6 months (excluding auto and mortgage inquiries) |
| | Total Revenue High Limit | Total revolving high credit limit |
| | Total account | The total number of credit lines currently in the borrower's credit file |
| | Finance inquiries | Number of personal finance inquiries |
| | credit age | How long has the earliest account been opened by the borrower |
| | Delinquencies | The number of delinquencies in the borrower's credit file for the past 2 years |
| | Public record | Number of derogatory public records |
| | Open account | The number of open credit lines in the borrower's credit file. |
| | Revolving balance | Total credit revolving balance |

Table 2. Correlation with respect to loan status

| NO | Attributes | Correlation |
|---|---|---|
| 1 | Lc Grade | -0.242 |
| 2 | Interest Rate | -0.224 |
| 3 | New Dti | -0.171 |
| 4 | Income to Payment Ratio | 0.141 |
| 5 | Dti | -0.132 |
| 6 | Home ownership RENT | -0.101 |
| 7 | Revolving utilization rate | -0.100 |
| 8 | percent_bc_gt_75 | -0.096 |
| 9 | Average Current Balance | 0.095 |
| 10 | Term | 0.094 |
| 11 | Home ownership MORTGAGE | 0.092 |
| 12 | Total Current Balance | 0.089 |
| 13 | Verification status Verified | -0.077 |
| 14 | installment | -0.076 |
| 15 | Verification status - not verified | 0.074 |
| 16 | inquiries last 6 months | -0.074 |
| 17 | Loan amount | -0.067 |
| 18 | Annual Income | 0.066 |
| 19 | Total Revenue High Limit | 0.064 |
| 20 | Employment Length | 0.043 |

A brief description of the final dataset appears in Table 3.

Table 3. Final dataset description

| N | Features | % Default | % Fully paid | Imbalance Ratio |
|---|---|---|---|---|
| 66376 | 43 | 18.3 | 81.7 | 4.46 |

"N" represents the number of instances in the dataset, "features" is the number of variables we used in the prediction analysis. "% default", "% fully paid" and "imbalance ratio" are reflective of "loan status" in the Lending Club dataset. Loans with a status of "current" have not yet been issued or have not reached maturity and, therefore, do not contain information a borrower's creditworthiness. These records were removed. "% default" reflects the proportion once the current loans had been removed with a status of "defaulted"; likewise, "% fully paid" reflects the proportion of loans that have reached maturity.

### 4.2. Performance Measurement

Accuracy is traditionally the most popular performance metric in a binary classification problem. However, when assessing imbalanced datasets, accuracy tends to emphasize the majority class, making it difficult for the classifier to perform well on the minority class. Moreover, in credit risk prediction, measuring accuracy does not consider that false positives are worse than false negatives and, therefore, accuracy can be a misleading criterion that causes erroneous results [28]. As such, other performance measures are more appropriate when working in domains with class imbalance issues [29]. Receiver operating characteristics (ROC), G-mean, and F-measure (FM) are preferred as the likelihood these measures will be affected by imbalanced class distributions is low [11].

In a binary classification problem with good and bad class labels, the classifier's result is considered successful if both the false positive rate and false negative rate are small [15]. Sensitivity measures the accuracy of positive samples and specificity measures negative samples. Additionally, an effective performance measure in imbalanced settings will indicate the balance between classification performance in both the minority and majority class [30]. The G-mean measure considers both sensitivity and specificity for both classes in calculating its scores and is therefore an effective measure for imbalanced datasets [15]. The G-mean equation is shown in Eq. (5).

$$G\text{-Mean} = \sqrt{Sensitivity * Specificity} \quad (5)$$

where

$$Sensivity = \frac{True\ positive}{True\ Positive + False\ Negative}$$

$$Specificity = \frac{True\ Negative}{True\ Negative + False\ positive}$$

The area under curve (AUC) measure determines the area under receiver operating characteristic (ROC) curve. This measure is another effective metric for measuring classification performance with imbalanced datasets [15].

### 4.3. Classification Results

In this empirical evaluation, we evaluated the performance of selected classifiers in combination with various resampling methods. The combinations are shown in Table 4.

The credit risk prediction results of the three selected classifiers – random forest, logistic regression, and linear discriminate analysis – were tested with each type of resampling method in groups. The best pair from each group was then compared with each other and with a non-sampling approach, i.e., with an imbalanced dataset that had not been resampled.

Table 4. Classifier / resampling method combinations

| Classifier | Resampling methodology | | Performance measure |
|---|---|---|---|
| Logistic regression | Under-sampling | RUS IHT | G-mean |
| Linear discriminant analysis | Over-sampling | ROS SMOTE ADASYN | AUC Sensitivity Specificity |
| Random forest | Hybrid | SMOTE-TOMEK SMOTE-ENN | False positive rate |

*For this and all subsequent tables: LR = logistic regression, LDA = linear discriminate analysis, RF = random forest, ROS = random over-sampling, RUS = random under-sampling, IHT = instance hardness threshold, SMOTE = synthetic minority over-sampling technique, ADASYN = adaptive synthetic sampling, EEN = edited nearest neighbor*

As shown in Table 5, the effectiveness of the different under-sampling methods depends on the measure used to gauge performance. As discussed in Section 4.2, G-means is the most effective measure for assessing the classification results when class imbalance problems exist in the dataset. From a G-means perspective, all classifiers in combination with RUS significantly outperformed the combinations with the instance hardness threshold method. The RF-RUS method had the highest G-mean (0.65). Therefore, RF-RUS is the best classification approach in the under-sampling group.

Table 5. Classification Results (Under-sampling approach)

| Classifier | Accuracy | AUC | Sensitivity | Specificity | FP-Rate | G-mean |
|---|---|---|---|---|---|---|
| RF-RUS | 0.692 | 0.69 | 0.717 | 0.582 | 0.42 | 0.65 |
| LR-RUS | 0.693 | 0.71 | 0.723 | 0.558 | 0.442 | 0.635 |
| LDA-RUS | 0.676 | 0.7034 | 0.695 | 0.589 | 0.42 | 0.64 |
| LR-IHT | 0.71 | 0.7 | 0.76 | 0.51 | 0.49 | 0.62 |
| LDA-IHT | 0.713 | 0.7 | 0.759 | 0.505 | 0.494 | 0.619 |
| RF-IHT | 0.75 | 0.688 | 0.83 | 0.4 | 0.61 | 0.57 |

The performance of the over-sampling methods with different classifiers appears in Table 6. In terms of accuracy, adaptive synthetic sampling behaved very differently when combined with a random forest than the other two classifiers. RF-ADASYN achieved an overall accuracy of 0.8 compared to 0.64 for LR-ADASYN and 0.61 for LDA-ADASYN. In addition, RF-ADASYN achieved the highest sensitivity (0.94); however, it fell to the bottom of the list for specificity. This demonstrates RF-ADASYN's poor performance in accurately predicting defaulters, and its bias toward the majority class despite an over-sampling approach designed to solve class imbalance issues. Random forest showed the lowest specificity and G-mean, but had the highest false positive rate of all the over-sampling techniques. These results indicate that random forest is an inappropriate classifier to hybridize with an over-sampling approach. When logistic regression and linear discriminate analysis were hybridized with over-sampling techniques, their AUC sat at around 0.7, whereas random forest fell to the bottom of the list in terms of AUC at between 0.65 and 0.68. Among all the over-sampling methods, LDA-SMOTE emerged as the best classification method according to the G-mean measure.

Table 6. Classification Result (Over-sampling approach)

| Classifier | Accuracy | AUC | Sensitivity | Specificity | FP-Rate | G-mean |
|---|---|---|---|---|---|---|
| LDA-SMOTE | 0.64 | 0.7 | 0.63 | 0.65 | 0.35 | 0.643 |
| LR-SMOTE | 0.6479 | 0.702 | 0.641 | 0.644 | 0.356 | 0.642 |
| LDA-ADASYN | 0.61 | 0.7 | 0.59 | 0.7 | 0.3 | 0.642 |
| LDA-ROS | 0.648 | 0.702 | 0.65 | 0.64 | 0.359 | 0.64 |
| LR-ADASYN | 0.64 | 0.7 | 0.64 | 0.64 | 0.36 | 0.64 |
| LR-ROS | 0.7 | 0.703 | 0.735 | 0.542 | 0.458 | 0.63 |
| RF-ROS | 0.699 | 0.689 | 0.74 | 0.513 | 0.487 | 0.616 |
| RF-SMOTE | 0.6814 | 0.658 | 0.725 | 0.486 | 0.513 | 0.594 |
| RF-ADASYN | 0.8 | 0.66 | 0.94 | 0.16 | 0.84 | 0.39 |

Given the performance measures in Table.7, the hybrid re-sampling methods did not produce good results. All methods had low accuracy. The models using SMOTE-ENN resulted in lower false positive rates and higher specificity, but their sensitivity scores were incredibly low. This is an indication that the model identifies most customers as defaulters based on significantly high performance in the minority class (the defaulting customers) while showing low performance on the other classes. For example, RF-SMOTE-ENN only correctly predicted 33.7% of the customers as good with a class of 1. Despite the poor performance of models presented in Table 7, LR-SMOTE-Tomek, with a G-means of 0.64, represents the best classification of the hybrid re-sampling methods.

Table 7. Classification Result (Hybrid approach)

| Classifier | Accuracy | AUC | Sensitivity | Specificity | FP-Rate | G-mean |
|---|---|---|---|---|---|---|
| LR-SMOTETomek | 0.64 | 0.7 | 0.638 | 0.648 | 0.352 | 0.643 |
| LDA-SMOTETomek | 0.64 | 0.701 | 0.637 | 0.646 | 0.354 | 0.642 |
| RF-SMOTETomek | 0.68 | 0.66 | 0.705 | 0.516 | 0.483 | 0.603 |
| LR-SMOTEENN | 0.47 | 0.699 | 0.377 | 0.862 | 0.138 | 0.57 |
| LDA-SMOTEENN | 0.46 | 0.698 | 0.37 | 0.86 | 0.137 | 0.566 |
| RF-SMOTEENN | 0.43 | 0.664 | 0.337 | 0.84 | 0.15 | 0.53 |

Next, we compared the best combinations from each group and list the results in Table 8. Table 8 also includes the results of the two non-sampling strategies, logistic regression and random forest, to illustrate the significant difference between prediction results on an imbalanced dataset versus a balanced one.

The non-sampling strategies were the most accurate; however, they performed worst in terms of G-mean, specificity, and false positive rates. This indicates that these classification models, by considering all the samples in the majority class (i.e., the good customers), are biased towards the majority class, but tend to ignore the minority.

Table 8. Classification Results (Final Comparison)

| Classifier | Accuracy | AUC | Sensitivity | Specificity | FP-Rate | G-mean |
|---|---|---|---|---|---|---|
| RF-RUS | 0.692 | 0.69 | 0.717 | 0.582 | 0.42 | 0.65 |
| LDA-SMOTE | 0.64 | 0.7 | 0.63 | 0.65 | 0.35 | 0.643 |
| LR-Tomek | 0.64 | 0.7 | 0.638 | 0.648 | 0.352 | 0.643 |
| Logistic Regression | 0.8173 | 0.703 | 0.988 | 0.048 | 0.95 | 0.218 |
| Random forest | 0.8176 | 0.696 | 0.996 | 0.015 | 0.98 | 0.12 |

In terms of G-mean, random forest had the lowest rating (0.12) on an imbalanced dataset. While, RF-RUS (i.e., random forest on a balanced dataset) had the highest G of 0.65. However, random forest and logistic regression returned the highest false positive rates on their own at 0.98 and 0.95, respectively, and 0.48 and 0.35, respectively, when combined with a resampling approach. This notable difference between non-sampling and resampling proves the effectiveness of resampling techniques on the performance of prediction modeling.

RF-RUS emerged as the best method for predicting a borrower's status in a social lending marketplace. Our computational results indicate that considering a random under-sampling technique for class imbalance issues, then use a random forest classifier on the resulting balanced training set outperforms the other methods examined in the experiments.

## 5. Conclusions and Future Works

Identifying the risk score for a potential borrower is crucial for the healthy functioning of social lending markets, where class imbalance problems are prevalent. However, few studies into social lending platforms have considered the characteristics of imbalanced data. Moreover, the efficiency of resampling techniques in evaluating P2P loans is a controversial issue.

To calculate the creditworthiness of borrowers in P2P lending platforms, we used the most recent data published by the Lending Club. Appropriate features were selected through comprehensive feature engineering process, and we introduced a non-standard financial feature to increase the reliability of the computed risk scores. Additionally, given the Lending Club dataset contains imbalanced classes, we also compared different resampling methods to determine the best overall technique. Accordingly, the state-of-the-art classifiers – random forest, logistic regression and linear discriminate analysis – were combined with different resampling techniques and tested on Lending Club's data. Our experiments show that random forest and random under-sampling may be an efficient combination of classifier and resampling strategy to compute risk scores for loan applicants in social lending markets.

P2P lenders can take advantage of the credit risk prediction modeling discussed in this study to make smarter decisions when evaluating loan applications. Moreover, lenders might apply the attributes identified in this study to compute the creditworthiness of borrowers. Identifying default borrowers in advance can prevent financial loss. Furthermore, more accurate assessments of the probability of default may also help when developing strategies to compensate for risk, such as increased interest rates.

One area for future research may consider the support vector machine as a classifier. A support vector machine algorithm may show better performance than the algorithms assessed in this paper, even though fine-tuning the parameters is time-consuming. Additionally, the Lending Club regularly publishes its historical data. Given the underlying distribution of incoming data may change unpredictably over time, i.e., the data may contain concept drift, these changes could affect the accuracy of prediction models in future. Therefore, another area for future work could focus on concept drift in imbalanced data streams on social lending platforms.

## References

1. Guo, Y., Zhou, W., Luo, C., Liu, C., and Xiong, H.: 'Instance-based credit risk assessment for investment decisions in P2P lending', European Journal of Operational Research, 2016, 249, (2), pp. 417-426
2. Malekipirbazari, M., and Aksakalli, V.: 'Risk assessment in social lending via random forests', Expert Systems with Applications, 2015, 42, (10), pp. 4621-4631
3. Emekter, R., Tu, Y., Jirasakuldech, B., and Lu, M.: 'Evaluating credit risk and loan performance in online Peer-to-Peer (P2P) lending', Applied Economics, 2015, 47, (1), pp. 54-70
4. Xia, Y., Liu, C., and Liu, N.: 'Cost-sensitive boosted tree for loan evaluation in peer-to-peer lending', Electronic Commerce Research and Applications, 2017, 24, pp. 30-49
5. Lin, W.-C., Tsai, C.-F., Hu, Y.-H., and Jhang, J.-S.: 'Clustering-based undersampling in class-imbalanced data', Information Sciences, 2017, 409, pp. 17-26
6. Ala'raj, M., and Abbod, M.F.: 'Classifiers consensus system approach for credit scoring', Knowledge-Based Systems, 2016, 104, pp. 89-105
7. Byanjankar, A., Heikkilä, M., and Mezei, J.: 'Predicting credit risk in peer-to-peer lending: A neural network approach', in Editor (Ed.)^(Eds.): 'Book Predicting credit risk in peer-to-peer lending: A neural network approach' (IEEE, 2015, edn.), pp. 719-725
8. Siami, M., Gholamian, M.R., Basiri, J., and Fathian, M.: 'An Application of Locally Linear Model Tree Algorithm for Predictive Accuracy of Credit Scoring', in Editor (Ed.)^(Eds.): 'Book An Application of Locally Linear Model Tree Algorithm for Predictive Accuracy of Credit Scoring' (Springer, 2011, edn.), pp. 133-142
9. Siami, M., Gholamian, M.R., and Basiri, J.: 'An application of locally linear model tree algorithm with combination of feature selection in credit scoring', International Journal of Systems Science, 2014, 45, (10), pp. 2213-2222
10. Chawla, N.V., Japkowicz, N., and Kotcz, A.: 'Special issue on learning from imbalanced data sets', ACM Sigkdd Explorations Newsletter, 2004, 6, (1), pp. 1-6
11. Haixiang, G., Yijing, L., Shang, J., Mingyun, G., Yuanyue, H., and Bing, G.: 'Learning from class-imbalanced data: review of methods and applications', Expert Systems with Applications, 2017, 73, pp. 220-239
12. Zhu, B., Baesens, B., Backiel, A., and vanden Broucke, S.K.: 'Benchmarking sampling techniques for imbalance learning in churn prediction', Journal of the Operational Research Society, 2017, pp. 1-17
13. Chawla, N.V., Bowyer, K.W., Hall, L.O., and Kegelmeyer, W.P.: 'SMOTE: synthetic minority over-sampling technique', Journal of artificial intelligence research, 2002, 16, pp. 321-357
14. He, H., Bai, Y., Garcia, E.A., and Li, S.: 'ADASYN: Adaptive synthetic sampling approach for imbalanced learning', in Editor (Ed.)^(Eds.): 'Book ADASYN: Adaptive synthetic sampling approach for imbalanced learning' (IEEE, 2008, edn.), pp. 1322-1328
15. Gong, J., and Kim, H.: 'RHSBoost: Improving classification performance in imbalance data', Computational Statistics & Data Analysis, 2017, 111, pp. 1-13
16. Smith, M.R., Martinez, T., and Giraud-Carrier, C.: 'An instance level analysis of data complexity', Machine learning, 2014, 95, (2), pp. 225-256

17. Abeysinghe, C., Li, J., and He, J.: 'A Classifier Hub for Imbalanced Financial Data', in Editor (Ed.)^(Eds.): 'Book A Classifier Hub for Imbalanced Financial Data' (Springer, 2016, edn.), pp. 476-479
18. Brown, I., and Mues, C.: 'An experimental comparison of classification algorithms for imbalanced credit scoring data sets', Expert Systems with Applications, 2012, 39, (3), pp. 3446-3453
19. Sanz, J.A., Bernardo, D., Herrera, F., Bustince, H., and Hagras, H.: 'A compact evolutionary interval-valued fuzzy rule-based classification system for the modeling and prediction of real-world financial applications with imbalanced data', IEEE Transactions on Fuzzy Systems, 2015, 23, (4), pp. 973-990
20. Liu, W., Wang, Z., Liu, X., Zeng, N., Liu, Y., and Alsaadi, F.E.: 'A survey of deep neural network architectures and their applications', Neurocomputing, 2017, 234, pp. 11-26
21. Koutanaei, F.N., Sajedi, H., and Khanbabaei, M.: 'A hybrid data mining model of feature selection algorithms and ensemble learning classifiers for credit scoring', Journal of Retailing and Consumer Services, 2015, 27, pp. 11-23
22. Umarani, J., and Manikandan, S.: 'Generating Enhanced Web Log File using Advanced Data Cleansing Algorithm in Pre-Processing Phase', Indian Journal of Science and Technology, 2016, 9, (48), pp. 1-7
23. de Oliveira, E.C., de Faro Orlando, A., dos Santos Ferreira, A.L., and de Oliveira Chaves, C.E.: 'Comparison of different approaches for detection and treatment of outliers in meter proving factors determination', Flow Measurement and Instrumentation, 2016, 48, pp. 29-35
24. Xia, Y., Liu, C., Li, Y., and Liu, N.: 'A boosted decision tree approach using Bayesian hyper-parameter optimization for credit scoring', Expert Systems with Applications, 2017, 78, pp. 225-241
25. James, G., Witten, D., Hastie, T., and Tibshirani, R.: 'An introduction to statistical learning' (Springer, 2013. 2013)
26. Kumar, K., and Bhattacharya, S.: 'Artificial neural network vs linear discriminant analysis in credit ratings forecast: A comparative study of prediction performances', Review of Accounting and Finance, 2006, 5, (3), pp. 216-227
27. Khemakhem, S., and Boujelbene, Y.: 'Credit risk prediction: A comparative study between discriminant analysis and the neural network approach', Accounting and Management Information Systems, 2015, 14, (1), pp. 60
28. Abellán, J., and Castellano, J.G.: 'A comparative study on base classifiers in ensemble methods for credit scoring', Expert Systems with Applications, 2017, 73, pp. 1-10
29. Galar, M., Fernandez, A., Barrenechea, E., Bustince, H., and Herrera, F.: 'A review on ensembles for the class imbalance problem: bagging-, boosting-, and hybrid-based approaches', IEEE Transactions on Systems, Man, and Cybernetics, Part C (Applications and Reviews), 2012, 42, (4), pp. 463-484
30. Wang, S., Minku, L.L., and Yao, X.: 'Online class imbalance learning and its applications in fault detection', International Journal of Computational Intelligence and Applications, 2013, 12, (04), pp.1-19